# An Algorithm for Computing Cusp Points in the Joint Space of *3-RPR* Parallel Manipulators

MAZEN ZEIN[†]    PHILIPPE WENGER [†]    DAMIEN CHABLAT[†]

**Abstract-** This paper presents an algorithm for detecting and computing the cusp points in the joint space of *3-RPR* planar parallel manipulators. In manipulator kinematics, cusp points are special points, which appear on the singular curves of the manipulators. The nonsingular change of assembly mode of *3-RPR* parallel manipulators was shown to be associated with the existence of cusp points. At each of these points, three direct kinematic solutions coincide. In the literature, a condition for the existence of three coincident direct kinematic solutions was established, but has never been exploited, because the algebra involved was too complicated to be solved. The algorithm presented in this paper solves this equation and detects all the cusp points in the joint space of these manipulators.

## I. Introduction

Critical points such as bifurcation points, turning points and cusp points can be considered of particular interest in applied sciences. The application of cusp points on manipulators kinematics is what we will be considering here.

A serial (resp. parallel) manipulator with cusp points in its workspace (resp. in its joint space) can change posture (resp. assembly mode) without crossing a singularity. When no cusp points exist, such a singular-free motion is not possible.

Regarding serial manipulators, it had been widely believed that they all should meet a singularity when changing posture. "Parenti-Castelli and Innocenti (1988)" were the first to prove the existence of non-singular posture changing motions on in 6-DOF serial manipulators. Similar results were found for 3R serial manipulators in "Burdick (1991)" and "Wenger (1992)". "Elomri and Wenger (1995)"

---

[†] Institut de Recherche en Communications et Cybernétique de Nantes UMR CNRS 6597

Correspondant author: Mazen.Zein@irccyn.ec-nantes.fr

have shown that serial manipulators having the ability to change posture without meeting a singularity are characterised by the existence of cusp points in their workspace, where three inverse kinematic solutions coincide. Cusp points in serial manipulators can be determined by looking for the triple roots of the inverse kinematics polynomial "Elomri and Wenger (1995)" or from the equation of the workspace boundary "Ottaviano and Husty (2004)".

Most fully parallel manipulators have multiple direct kinematic solutions, which are associated with the different assembly modes. "Hunt and Primrose (1993)" first showed that to move from one assembly mode to another, a fully parallel manipulator had to cross a singularity. But, "Innocenti and Parenti-Castelli (1998)" found a *3-RPR* parallel manipulator able to change its assembly mode without crossing a singularity. One year later, "Mcaree (1999)" pointed out that *3-RPR* and octahedral manipulators can undertake non-singular assembly changing motions, if a point with triple direct kinematic solutions exists in their joint space, this point is "a cusp point" in a section of the joint space. He established a condition for the existence of cusp points. But this condition has never been exploited, because the algebra involved in this condition was found to be too complicated. "Wenger and Chablat (1998)" showed that to accomplish a non-singular assembly-mode changing motion, a *3-RPR* manipulator platform should encircle a cusp point in its joint space. Thus, the determination of the cusp points is of interest for planning trajectories.

In this paper, an algorithm for detecting all cusp points and computing their coordinates in the joint space of *3-RPR* parallel manipulators is established; it is based on the abovementioned condition. This work finds application in both design and trajectory planning.

In the following sections, we present the *3-RPR* parallel manipulators studied and their constraint equations, we explain briefly the cusp points existence condition established by "Mcaree (1999)", then the algorithm is described and run on two different *3-RPR* manipulators.

## II. Preliminaries

*II.1 Manipulators studied*

The manipulators under study are 3-DOF planar parallel manipulators with three extensible leg rods (Fig. 1). These manipulators have been frequently studied, for example by "Sefrioui and Gosselin (1995)" and "Merlet (2000)". Each of the three extensible leg rods is actuated with a prismatic joint. The geometric parameters of the manipulators are the three sides of the moving platform $d_1$, $d_2$, $d_3$ and the position of the base revolute joint centres defined by $A_1$, $A_2$ and $A_3$. The reference frame is centred at $A_1$ and the *x*-axis passes through $A_2$. Thus, $A_1 = (0, 0)$, $A_2 = (A_{2x}, 0)$ and $A_3 = (A_{3x}, A_{3y})$.

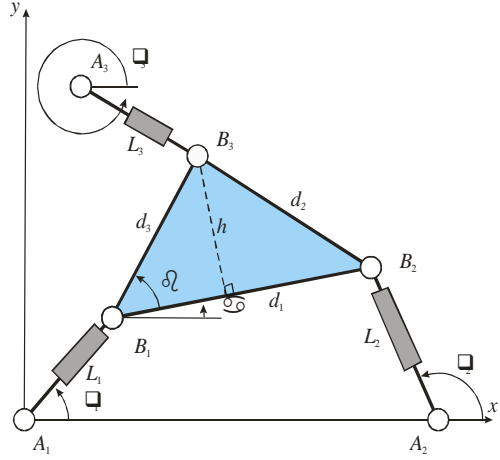

Fig 1. The 3-R<u>P</u>R parallel manipulator under study.

*II.2 Constraint equations*

Let $\mathbf{L} \equiv (L_1, L_2, L_3)$ define the lengths of the three leg rods and let $\boldsymbol{\theta} \equiv (\theta_1, \theta_2, \theta_3)$ define the three angles between the leg rods and the *x*-axis. The six parameters $(\mathbf{L}, \boldsymbol{\theta})$ can be regarded as a configuration of the manipulator but only three of them are independent, so that the configuration space is a 3-dimensional manifold embedded in a 6-dimensional space. The dependency between $(\mathbf{L}, \boldsymbol{\theta})$ can be identified by writing the fixed distances between the three vertices of the mobile platform $B_1$, $B_2$, $B_3$, which yield the following constraint equations

$$\begin{cases} \Gamma_1(\mathbf{L},\boldsymbol{\theta}) = [\mathbf{b}_2(\mathbf{L},\boldsymbol{\theta}) - \mathbf{b}_1(\mathbf{L},\boldsymbol{\theta})]^T [\mathbf{b}_2(\mathbf{L},\boldsymbol{\theta}) - \mathbf{b}_1(\mathbf{L},\boldsymbol{\theta})] - d_1^2 = 0 \\ \Gamma_2(\mathbf{L},\boldsymbol{\theta}) = [\mathbf{b}_3(\mathbf{L},\boldsymbol{\theta}) - \mathbf{b}_2(\mathbf{L},\boldsymbol{\theta})]^T [\mathbf{b}_3(\mathbf{L},\boldsymbol{\theta}) - \mathbf{b}_2(\mathbf{L},\boldsymbol{\theta})] - d_2^2 = 0 \\ \Gamma_3(\mathbf{L},\boldsymbol{\theta}) = [\mathbf{b}_1(\mathbf{L},\boldsymbol{\theta}) - \mathbf{b}_3(\mathbf{L},\boldsymbol{\theta})]^T [\mathbf{b}_1(\mathbf{L},\boldsymbol{\theta}) - \mathbf{b}_3(\mathbf{L},\boldsymbol{\theta})] - d_3^2 = 0 \end{cases} \quad (1)$$

where $\mathbf{b}_i$ is the vector defining the coordinates of $B_i$ in the reference frame as function of $\mathbf{L}$ and $\boldsymbol{\theta}$. For more simplicity, $(\mathbf{L}, \boldsymbol{\theta})$ will be omitted in the following equations.

Expanding each $\Gamma_i$ as a series about the configuration $(\mathbf{L}, \boldsymbol{\theta})$ yields

$$\Delta\Gamma_i = \left( \sum_{j=1}^{3} \Delta\theta_i \frac{\partial}{\partial\theta_j} + \sum_{j=1}^{3} \Delta L_i \frac{\partial}{\partial L_j} \right) \Gamma_i + \frac{1}{2!}\left( \sum_{j=1}^{3} \Delta\theta_i \frac{\partial}{\partial\theta_j} + \sum_{j=1}^{3} \Delta L_i \frac{\partial}{\partial L_j} \right)^2 \Gamma_i + ... + \\ \frac{1}{n!}\left( \sum_{j=1}^{3} \Delta\theta_i \frac{\partial}{\partial\theta_j} + \sum_{j=1}^{3} \Delta L_i \frac{\partial}{\partial L_j} \right)^n \Gamma_i + .. = 0 \quad (2)$$

If one keeps only the first-order and second-order terms, Eq. (2) can be written in matrix form as follows

$$\Delta\Gamma = \frac{\partial\Gamma}{\partial\boldsymbol{\theta}}\Delta\boldsymbol{\theta} + \frac{\partial\Gamma}{\partial\mathbf{L}}\Delta\mathbf{L} + \frac{1}{2}\begin{bmatrix} \Delta\boldsymbol{\theta}^T \frac{\partial^2\Gamma_1}{\partial\boldsymbol{\theta}^2}\Delta\boldsymbol{\theta} \\ \Delta\boldsymbol{\theta}^T \frac{\partial^2\Gamma_2}{\partial\boldsymbol{\theta}^2}\Delta\boldsymbol{\theta} \\ \Delta\boldsymbol{\theta}^T \frac{\partial^2\Gamma_3}{\partial\boldsymbol{\theta}^2}\Delta\boldsymbol{\theta} \end{bmatrix} + \begin{bmatrix} \Delta\boldsymbol{\theta}^T \frac{\partial^2\Gamma_1}{\partial\boldsymbol{\theta}\partial\mathbf{L}}\Delta\mathbf{L} \\ \Delta\boldsymbol{\theta}^T \frac{\partial^2\Gamma_2}{\partial\boldsymbol{\theta}\partial\mathbf{L}}\Delta\mathbf{L} \\ \Delta\boldsymbol{\theta}^T \frac{\partial^2\Gamma_3}{\partial\boldsymbol{\theta}\partial\mathbf{L}}\Delta\mathbf{L} \end{bmatrix} + \frac{1}{2}\begin{bmatrix} \Delta\mathbf{L}^T \frac{\partial^2\Gamma_1}{\partial\mathbf{L}^2}\Delta\mathbf{L} \\ \Delta\mathbf{L}^T \frac{\partial^2\Gamma_2}{\partial\mathbf{L}^2}\Delta\mathbf{L} \\ \Delta\mathbf{L}^T \frac{\partial^2\Gamma_3}{\partial\mathbf{L}^2}\Delta\mathbf{L} \end{bmatrix} = 0 \quad (3)$$

Equation (3) can be used to describe an arbitrary local motion at a given configuration of the manipulator "Mcaree (1999)". When first order terms of Eq. (3) are sufficient to describe the motion, the manipulator is in a regular configuration and the following equation can be used instead of Eq. (3)

$$\frac{\partial \Gamma(\mathbf{L},\boldsymbol{\theta})}{\partial \boldsymbol{\theta}}\Delta\boldsymbol{\theta}+\frac{\partial \Gamma(\mathbf{L},\boldsymbol{\theta})}{\partial \mathbf{L}}\Delta\mathbf{L}=0 \quad (4)$$

Otherwise the configuration (**L**, θ) is special and the manipulator meets a singularity. This happens when the constraint Jacobian ∂Γ/∂θ drops rank so that the second order terms of the equation (3) are needed to describe the constraints. The three vertices of the moving platform have the following coordinates in the fixed reference frame

$$\mathbf{b}_1 = \begin{bmatrix} L_1 \cos(\theta_1) & L_1 \sin(\theta_1) \end{bmatrix}^T ; \mathbf{b}_2 = \begin{bmatrix} A_{2x} + L_2 \cos(\theta_2) & L_2 \sin(\theta_2) \end{bmatrix}^T ;$$
$$\mathbf{b}_3 = \begin{bmatrix} A_{3x} + L_3 \cos(\theta_3) & A_{3y} + L_3 \sin(\theta_3) \end{bmatrix}^T .$$

Thus, the constraint Jacobian can be put in the following form

$$\frac{\partial \Gamma}{\partial \boldsymbol{\theta}} = 2 \begin{bmatrix} L_1(A_{2x}s_1 + L_2 s_{12}) & L_2(L_1 s_{21} - A_{2x}s_2) & 0 \\ 0 & \begin{matrix}-L_2((A_{2x}-A_{3x})s_2 \\ -L_3 s_{23} + A_{3y}c_2)\end{matrix} & \begin{matrix}L_3((A_{2x}-A_{3x})s_3 \\ -L_2 s_{23} + A_{3y}c_3)\end{matrix} \\ \begin{matrix}L_1(A_{3x}s_1 - L_3 s_{31} \\ -A_{3y}c_1)\end{matrix} & 0 & \begin{matrix}-L_3(A_{3x}s_3 - L_1 s_{31} \\ -A_{3y}c_3)\end{matrix} \end{bmatrix} \quad (5)$$

where $s_i = \sin(\theta_i)$, $c_i = \cos(\theta_i)$ and $s_{ij} = \sin(\theta_i - \theta_j)$.

## III. Existence condition of cusp points

To determine the cusp points, we need first to characterise the singular configurations. "Mcaree (1999)" determined the singularities of the *3-RPR* manipulator by looking for the configurations where det(∂Γ/∂θ) vanishes. In our work, we have used a geometric approach that is much more direct than the calculation of the determinant, which does not simplify easily. It is well known that a 3-RPR manipulator is in a singular configuration whenever the axes of its three leg rods intersect (possibly at infinity). The derivation of this geometric condition is straightforward and yields the following equation

$$A_{2x}s_2 s_{31} + (A_{3x}s_3 - A_{3y}c_3)s_{12} = 0 \quad (6)$$

For serial 3-DOF manipulators, the cusp points can be determined by deriving the condition under which the inverse kinematics polynomial admits three identical roots (Elomri 1995). However this approach is much more complicated when applied to the direct kinematics polynomial of 3-*RPR* manipulators because this polynomial is of degree 6.

An interesting alternative approach was proposed in "Mcaree (1999)" by writing the condition under which the manipulator loses first and second order constraints. The resulting condition for triple coalescence of assembly modes was shown to take the following form

$$\mathbf{v}^T \left[ u_1 \frac{\partial^2 \Gamma_1}{\partial \boldsymbol{\theta}^2} + u_2 \frac{\partial^2 \Gamma_2}{\partial \boldsymbol{\theta}^2} + u_3 \frac{\partial^2 \Gamma_3}{\partial \boldsymbol{\theta}^2} \right] \mathbf{v} = 0 \quad (7)$$

where **v** is a unit vector in the right kernel of matrix $\partial\Gamma/\partial\theta$, and $u_1$, $u_2$, $u_3$ are the three components of the unit vector **u** that spans the left kernel. Vectors **u** and **v** can be chosen in the set of nonzero rows and columns of the adjoint of matrix $\partial\Gamma/\partial\theta$ (i.e. the matrix of cofactors of the transpose of $\partial\Gamma/\partial\theta$), respectively.

Calculating the adjoint of $\partial\Gamma/\partial\theta$ from Eq. (5) yields

$$\text{adj}\left(\frac{\partial\Gamma}{\partial\theta}\right) = \begin{pmatrix} k_1k_2 & -k_2k_5 & k_3k_5 \\ k_3k_4 & k_2k_6 & k_3k_6 \\ -k_1k_4 & k_4k_5 & k_1k_6 \end{pmatrix} \tag{8}$$

where

$$\begin{aligned}
k_1 &= 2L_2\left((A_{3x} - A_{2x})s_2 + L_3s_{23} - A_{3y}c_2\right) & k_4 &= 2L_1\left(L_3s_{13} + A_{3x}s_1 - A_{3y}c_1\right) \\
k_2 &= -2L_3\left(L_1s_{13} + A_{3x}s_3 - A_{3y}c_3\right) & k_5 &= -2L_2\left(L_1s_{12} + A_{2x}s_2\right) \\
k_3 &= -2L_3\left((A_{3x} - A_{2x})s_3 + L_2s_{23} - A_{3y}c_3\right) & k_6 &= 2L_1\left(L_2s_{12} + A_{2x}s_1\right)
\end{aligned} \tag{9}$$

Taking **u** (resp. **v**) as the first row (resp. column) of (9), the equation (7) can be written as

$$\begin{pmatrix} k_1k_2 & k_3k_4 & -k_1k_4 \end{pmatrix} \left( k_1k_2 \frac{\partial^2\Gamma_1}{\partial\theta^2} - k_2k_5 \frac{\partial^2\Gamma_2}{\partial\theta^2} + k_3k_5 \frac{\partial^2\Gamma_3}{\partial\theta^2} \right) \begin{pmatrix} k_1k_2 \\ k_3k_4 \\ -k_1k_4 \end{pmatrix} = 0 \tag{10}$$

where

$$\frac{\partial^2\Gamma_1}{\partial\theta^2} = 2\begin{bmatrix} L_1(A_{2x}c_1 + L_2c_{21}) & -L_1L_2c_{21} & 0 \\ -L_1L_2c_{21} & -L_2(A_{2x}c_2 - L_1c_{21}) & 0 \\ 0 & 0 & 0 \end{bmatrix}$$

$$\frac{\partial^2\Gamma_2}{\partial\theta^2} = 2\begin{bmatrix} 0 & 0 & 0 \\ 0 & \begin{matrix}-L_2((A_{2x}-A_{3x})c_2\\-L_3c_{23}-A_{3y}s_2)\end{matrix} & -L_2L_3c_{23} \\ 0 & -L_2L_3c_{23} & \begin{matrix}L_3((A_{2x}-A_{3x})c_3\\+L_2c_{23}-A_{3y}s_3)\end{matrix} \end{bmatrix} \tag{11}$$

$$\frac{\partial^2\Gamma_3}{\partial\theta^2} = 2\begin{bmatrix} L_1(A_{3x}c_1 + L_3c_{31} + A_{3y}s_1) & 0 & -L_1L_3c_{31} \\ 0 & 0 & 0 \\ -L_1L_3c_{31} & 0 & L_3(L_1c_{31} - A_{3x}c_3 - A_{3y}s_3) \end{bmatrix}$$

"Mcaree (1999)" left equation (10) as such and no information was provided on how to use it in a computer program. He noted that the expansion of this equation was too complicated to yield any real insight.

We have developed an algorithm to solve this equation for any 3-*RPR* manipulator and we have implemented it in Maple. This algorithm detects all the cusp points inside the joint space of any 3-*RPR* manipulators and computes their coordinates.

We present it in the next section, and we run it on two different *3-RPR* manipulators.

## IV. Algorithm for calculating cusp points

The existence of cusp points allows the *3-RPR* manipulator to undertake non-singular assembly mode changing trajectories, these special trajectories can be executed by encircling a cusp point. "Mcaree (1999)" stated that cusp points are pernicious and should be avoided or designed out by judicious dimensioning.

The configuration of the *3-RPR* manipulator is given by six parameters: the three rod lengths ($L_1$, $L_2$, $L_3$), and the platform position variables ($\theta_1$, $\theta_2$, $\theta_3$). Only three of these parameters are independent. In order to reduce the dimension of our problem, "Mcaree (1999)" shows that it is possible to consider two-dimensional slices of the configuration space by fixing one of the leg rod lengths.

By doing so, the manipulator configuration can be fully defined by only two parameters. For example, for a fixed value of $L_1$, a configuration may be fully defined by either ($\alpha, \theta_1$) or ($L_2$, $L_3$). Note that in the first case, the configuration is defined in the output space by the position and the orientation of the moving platform ($L_1$ and $\theta_1$ define the position of $B_1$ in the plane and $\alpha$ defines the orientation of the moving platform in the plane). In the second case, the configuration is defined in the joint space by the three leg rod lengths.

In our work, we have always taken $L_1$ as the fixed parameter. After fixing the value of $L_1$, we first calculate the singularity curves in ($L_2$, $L_3$), and then we compute all the cusp points of this two-dimensional slice.

*IV.1  Algorithm*

If we consider equation (6), we notice that it is a function of ($\theta_1$, $\theta_2$, $\theta_3$). The existence condition of cusp points (10) is a function of ($L_1$, $L_2$, $L_3$) and ($\theta_1$, $\theta_2$, $\theta_3$). Our first goal is to establish an equation, which is a function of ($L_1$, $\alpha$, $\theta_1$), and then to solve it to obtain the cusp points coordinates. Thus, we first consider the following set of equations computed from the geometry of the manipulator.

$$\begin{cases} \cos(\theta_2) = \dfrac{-A_{2x} + L_1 \cos(\theta_1) + d_1 \cos(\alpha)}{L_2} \\ \cos(\theta_3) = \dfrac{-A_{3x} + L_1 \cos(\theta_1) + d_3 \cos(\alpha + \beta)}{L_3} \\ \sin(\theta_2) = \dfrac{L_1 \sin(\theta_1) + d_1 \sin(\alpha)}{L_2} \\ \sin(\theta_3) = \dfrac{-A_{3y} + L_1 \sin(\theta_1) + d_3 \sin(\alpha + \beta)}{L_3} \end{cases} \quad (12)$$

The algorithm for detecting cusp points is implemented in MAPLE; its steps are presented below:

1. First, the expression of $\cos(\theta_2)$, $\cos(\theta_3)$, $\sin(\theta_2)$ and $\sin(\theta_3)$ in (12) are substituted into the singularities equations (6). Then, $\sin(\alpha)$, $\cos(\alpha)$, $\cos(\theta_1)$ and $\sin(\theta_1)$ are replaced by the tangents of their half angles $\tan(\alpha/2)$ and $\tan(\theta_1/2)$, as a consequence we obtain an equation of the form:

$$F_1(L_1, t, t_1) = 0 \quad (13)$$

   where $t = \tan(\alpha)$ and $t_1 = \tan(\theta_1)$.

2. Then, the expression of $\cos(\theta_2)$, $\cos(\theta_3)$, $\sin(\theta_2)$ and $\sin(\theta_3)$ in (12) are substituted into equations (9) and (11) and $\sin(\alpha)$, $\cos(\alpha)$, $\cos(\theta_1)$ and $\sin(\theta_1)$ are replaced by the tangents of their half angles $\tan(\alpha/2)$ and $\tan(\theta_1/2)$. We get an equation of the form:

$$E_1(L_1, t, t_1) = 0 \quad (14)$$

So, we notice that the two equations (6) and (10) are written now as function of three parameters only.

3. We fix now $L_1$, and we input the manipulator parameters $d_1$, $d_2$, $d_3$, $A_{2x}$, $A_{3x}$ and $A_{3y}$. We have noticed that the direct substitution of the real values of $\sin(\beta)$ and $\cos(\beta)$ into the equations (13) and (14) make the equations resolution very complicated in the following steps. Thus, we write $\sin(\beta)$ and $\cos(\beta)$ as a function of an intermediate parameter $h$, which is the altitude of the moving triangle.

4. The Maple *resultant* function is used to eliminate $t = \tan(\alpha)$ from the two equations (13) and (14). The resulting equation is a polynomial of degree 96 in $t_1 = \tan(\theta_1)$, which can be factored as follows:

$$P_1^{a_1} P_2^{a_2} P_3^{a_3} ... P_{n-1}^{a_{n-1}} P_n^{a_n} Q = 0 \tag{15}$$

where $Q$ is a $24^{th}$-order univariate polynomial in $t_1$ and $P_1, P_2, \ldots, P_n$ are quadratic and quartic polynomials in $t_1$. Note that the factor form cannot be obtained without the intermediate parameter $h$.

5. We input the parameter $h$ value. We solve equation (15). Each real root $t_{1i}$ is back-substituted into (13), which is then solved for $t$. For every $t_{1i}$, we obtain different $t_{ij}$. Finally, we get a number of solution couples $(t_{ij}, t_{1i})$.

6. We substitute the values of each solution couple $(t_{ij}, t_{1i})$ into (14), and we keep only those that satisfy this equation.

7. The solutions $(t_{ij}, t_{1i})$ kept in the last step should give the coordinates of the cusp points. To verify this, we calculate the direct kinematic solutions for each solution $(t_{ij}, t_{1i})$. In many instances, we have found that some solutions do not yield three coincident solutions, which means that they are not associated with cusp points. So we reject them and we keep only those solutions that give three coincident direct kinematic solutions. These couples are the coordinates of the cusp points, we call them $(\alpha_{ij}, \theta_{1i})_{cusp}$.

After executing our algorithm hundreds of times, we have noticed that in each case all cusp points were determined by the $24^{th}$-order polynomial $Q$ of equation (15), that is, all remaining factors provided spurious solutions. Thus, we may conjecture that the cusp points are determined by $Q$, although we have no mathematical proof for this fact. All the real roots of $Q$ are the cusp points. With this conjecture, our algorithm simplifies significantly because instead of solving (16) (a $96^{th}$ order polynomial) we just have to solve polynomial $Q$ (a $24^{th}$ order polynomial).

To implement this result in the algorithm, we must change steps 5 and 6 into the following steps 5' and 6', and eliminate step 7:

5'. We input the parameter $h$ real value. We solve the polynomial $Q$. We substitute every real root $\theta_{1i}$ of $Q$ into equation (13), and we solve it for $\tan(\alpha/2)$. For every $\theta_{1i}$, we obtain different values $\alpha_{ij}$. Finally, we get a number of couples $(\alpha_{ij}, \theta_{1i})$.

6'. We substitute the values of each couple $(\alpha_{ij}, \theta_{1i})$ into equation (14). The couples that satisfy this equation are the cusp points coordinates. We call them $(\alpha_{ij}, \theta_{1i})_{cusp}$.

Finally, to obtain the coordinates of the cusp points in the joint space $(L_1, L_2, L_3)$, we use the following equations computed from the geometry of the manipulator:

$$L_2 = \sqrt{\left(-A_{2x} + L_1 \cos(\theta_1) + b_1 \cos(\alpha)\right)^2 + \left(L_1 \sin(\theta_1) + b_1 \sin(\alpha)\right)^2} \tag{17}$$

$$L_3 = \sqrt{\begin{aligned}&\left(A_{3x} - L_1 \cos(\theta_1) - b_3 \left(\cos(\alpha)\cos(\beta) - \sin(\alpha)\sin(\beta)\right)\right)^2 \\ &+ \left(A_{3y} - L_1 \sin(\theta_1) - b_3 \left(\sin(\alpha)\cos(\beta) - \cos(\alpha)\sin(\beta)\right)\right)^2\end{aligned}} \tag{18}$$

and we obtain the cusp points in a slice of the joint space for a fixed value of $L_1$.

*IV.2   Important conclusion*

In step 7 of the algorithm, we have noticed that the cusp existence condition generates solutions that do not provide triple direct kinematic solutions. This means that the cusp existence condition established by "Mcaree (1999)" is not a necessary and sufficient condition but only a sufficient condition.

*IV.3   Algorithm execution*

In this paragraph, we present the results of some executions of the algorithm for two different *3-RPR* parallel manipulators. Only some slices of the joint space are presented. However, for both manipulators, we have run the algorithm for $L_1$ varying from 0 to 50 with a scanning step of 0.1. We have noticed that the number of cusp points varies from one slice to another. Note that in serial manipulators, the number of cusp points does not depend on the workspace section (if we consider just the sections which passes through the axis of the first revolute joint).

On the other hand, the number of cusp points stabilizes for sufficiently large values of $L_1$. For example, there are always four cusp points for the first manipulators as soon as $L_1>31$.

Finally, the maximal number of cusp points depends on the geometry of the manipulator. For example, the second manipulator may have at most 6 cusp points whereas the first one may has 8 cusp points. We have also found manipulators with only 0, 2 or 4 cusp points. A symmetric manipulator with two similar platform has 0 cusp points as this manipulator is non-cuspidal "Mcaree (1999)". We have not been able to find manipulators with more than 8 cusp points.

The algorithm execution time slightly depends on the value of $L_1$ and of the *3-RPR* manipulator parameters. It highly depends on the number of digits required for the calculation. For 90 digits (which is necessary to guaranty a good accuracy), it is about two minutes on a computer equipped with a 3GHz-Pentium 4 with 512 Mo of Ram.

### IV.3.1   *3-RPR* parallel manipulator used in "Mcaree (1999)"

First, we begin with the *3-RPR* parallel manipulator used in "Mcaree (1999)" and "Innocenti (1998)". The geometric parameters of this manipulator are recalled below in an arbitrary length unit:

$A_1=(0, 0)$        $d_1=17.04$
$A_2=(15.91, 0)$    $d_2=16.54$
$A_3=(0, 10)$       $d_3=20.84$

➢ *A slice for $L_1=14.98$*

For the same fixed value $L_1=14.98$, as in "Mcaree (1999)" and "Innocenti (1998)", the algorithm detects six cusp points instead of five identified in "Mcaree (1999)". Figure 2 shows the singular curves of the 3-*RPR* manipulator for $L_1=14.98$, and the six cusp points pinpointed with circles. The sixth point missed by "Mcaree (1999)" is the point A, it is circled with bold lines and in-boxed in a separate view. The zoomed view shows that it is really a cusp point.

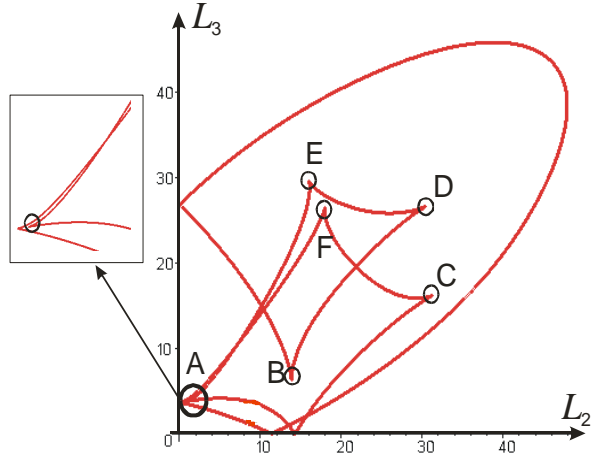

*Fig 2. Singular curves and cusp points in a slice of the 3-RPR manipulator joint space ($L_2$,$L_3$) for $L_1$=14.98.*

The coordinates of the six cusp points are given in the table 1 below

|  | $\alpha$ | $\theta_1$ | $L_2$ | $L_3$ |
|---|---|---|---|---|
| Cusp A | 50.67 deg | -69.12 deg | 0.84 | 3.77 |
| Cusp B | -2.59 deg | 177.32 deg | 13.85 | 6.26 |
| Cusp C | -122.89 deg | 114.05 deg | 31.27 | 16.17 |
| Cusp D | 57.48 deg | 133.77 deg | 30.44 | 26.61 |
| Cusp E | -0.59 deg | 15.46 deg | 16.02 | 29.56 |
| Cusp F | 170.37 deg | -10.65 deg | 17.98 | 26.44 |

*Tab 1. Coordinates of the six cusp points for $L_1$=14.98.*

➢ *A slice for $L_1$=34*

For the same manipulator with $L_1$=34, four cusp points are found. Figure 3 shows the singularity curves and the four cusps in the slice of the joint space for $L_1$=34.

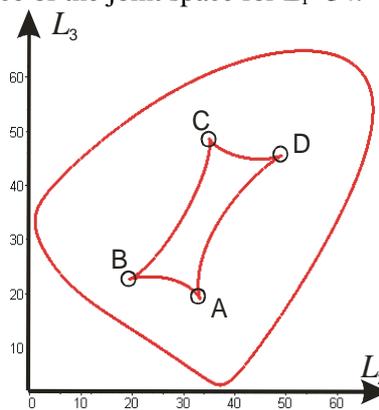

*Fig 3. Singular curves and cusp points in a slice of the 3-RPR manipulator joint space ($L_2$,$L_3$) for $L_1$=34.*

The coordinates of the four cusp points are given in the table 2 below

|  | α | θ₁ | $L_2$ | $L_3$ |
|---|---|---|---|---|
| Cusp A | -3.84 deg | -167.01 deg | 33.22 | 19.00 |
| Cusp B | 52.71 deg | -61.76 deg | 19.46 | 22.68 |
| Cusp C | -1.07 deg | 15.43 deg | 35.00 | 48.64 |
| Cusp D | 55.85 deg | 128.19 deg | 49.14 | 45.52 |

Tab 2.  Coordinates of the four cusp points for $L_1=34$.

➤ *A slice for $L_1=27$*

For the same manipulator with $L_1=27$, eight cusp points are found. Figure 4 shows the singularity curves and the eight cusps in the slice of the joint space for $L_1=27$.

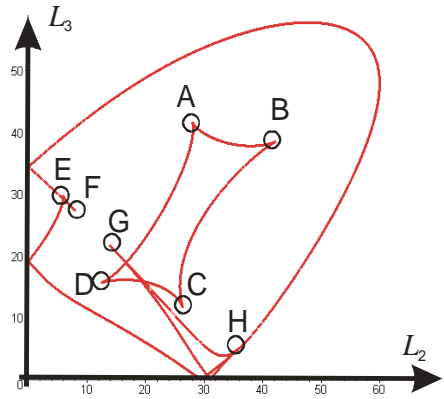

*Fig 4.  Singular curves and cusp points in a slice of the 3-RPR manipulator joint space ($L_2$, $L_3$) for $L_1=27$.*

The coordinates of the eight cusp points are given in the table 3 below

|  | α | θ₁ | $L_2$ | $L_3$ |
|---|---|---|---|---|
| Cusp A | -0.95 deg | 15.47 deg | 28.01 | 41.63 |
| Cusp B | 56.20 deg | 129.36 deg | 42.21 | 38.54 |
| Cusp C | -5.11 deg | -168.45 deg | 26.31 | 11.84 |
| Cusp D | 52.23 deg | -63.22 deg | 12.56 | 15.71 |
| Cusp E | -168.17 deg | 8.70 deg | 5.92 | 29.74 |
| Cusp F | -125.54 deg | 43.86 deg | 7.98 | 27.36 |
| Cusp G | -113.95 deg | 63.88 deg | 13.96 | 21.66 |
| Cusp H | -129.36 deg | 103.65 deg | 35.57 | 4.80 |

Tab 3.  Coordinates of the eight cusp points for $L_1=27$.

### IV.3.2  Another *3-RPR* parallel manipulator

The geometric parameters of the second manipulator are given below in an arbitrary length unit:

$A_1=(0, 0)$     $d_1=13$
$A_2=(30, 0)$    $d_2=9$
$A_3=(11, 27)$   $d_3=4$

➢ *For $L_1=3$:*

In this manipulator joint space section, the algorithm detects four cusp points. Figure 5 shows the singularity curves and the four cusps in the slice of the joint space for $L_1=3$.

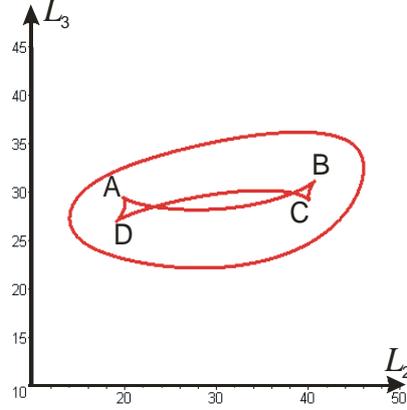

*Fig 5. Singular curves and cusp points in a slice of the joint space ($L_2,L_3$) of the second 3-RPR manipulator studied for $L_1=3$.*

The coordinates of the four cusp points are given in the table 4 below

|        | $\alpha$     | $\theta_1$   | $L_2$ | $L_3$ |
|--------|--------------|--------------|-------|-------|
| Cusp A | 12.52 deg    | -145.11 deg  | 19.80 | 29.43 |
| Cusp B | 156.76 deg   | -63.48 deg   | 40.68 | 31.11 |
| Cusp C | -168.74 deg  | 26.38 deg    | 40.08 | 29.14 |
| Cusp D | -20.32 deg   | 114.01 deg   | 19.11 | 27.01 |

*Tab 4. Coordinates of the four cusp points of the second manipulator studied for $L_1=3$.*

## V. Conclusions

In this paper, we have reviewed and exploited the cusp points existence condition defined by "Mcaree (1999)", we have found that it is only a sufficient condition and not a necessary one.

An algorithm, able to detect and to compute cusp points inside any section of the joint space of any *3-RPR* parallel manipulator, has been established. To the best of the authors' knowledge, such an algorithm had never been proposed before. The algorithm results in a $96^{th}$ degree univariate polynomial that can be put in a factored form. We have showed with intensive numerical experiments that the cusp points coordinates are the real roots of a $24^{th}$ degree univariate polynomial, which is one of the factors of the $96^{th}$ polynomial.

Finally, the results of four numerical executions of the algorithm on two different manipulators has been exposed. Determination of the cusp points is an important issue for planning non-singular assembly mode changing trajectories in parallel manipulators.

Contrary to serial manipulators, the number of cusp points is not the same in all sections. Future work will investigate the transition slices of the joint space where the number of cusp points changes, and its physical meaning for the behaviour of the manipulator.